\def\year{2020}\relax
\documentclass[letterpaper]{article} % DO NOT CHANGE THIS
\usepackage{aaai20}  % DO NOT CHANGE THIS
\usepackage{times}  % DO NOT CHANGE THIS
\usepackage{helvet} % DO NOT CHANGE THIS
\usepackage{courier}  % DO NOT CHANGE THIS
\usepackage[hyphens]{url}  % DO NOT CHANGE THIS
\usepackage{graphicx} % DO NOT CHANGE THIS
\usepackage{graphicx}  % DO NOT CHANGE THIS
\usepackage{amsmath}
\usepackage{amssymb}
\usepackage{multirow}
\usepackage{booktabs}
\usepackage{array}
\usepackage{algorithm}
\usepackage{algorithmic}
\usepackage{longtable}
\usepackage{enumitem}
\usepackage{enumerate}
\usepackage{bm}

\title{Discriminative Sentence Modeling for Story Ending Prediction }
\author{Yiming Cui\textsuperscript{\rm 1,2}, Wanxiang Che\textsuperscript{\rm 1}, Wei-Nan Zhang\textsuperscript{\rm 1}, Ting Liu\textsuperscript{\rm 1}, Shijin Wang\textsuperscript{\rm 2,3}, Guoping Hu\textsuperscript{\rm 2} \\ % All authors must be in the same font size and format. Use \Large and \textbf to achieve this result when breaking a line
\textsuperscript{\rm 1}Research Center for Social Computing and Information Retrieval (SCIR), Harbin Institute of Technology, Harbin, China\\ %If you have multiple authors and multiple affiliations
\textsuperscript{\rm 2}State Key Laboratory of Cognitive Intelligence, iFLYTEK Research, China\\
\textsuperscript{\rm 3}iFLYTEK AI Research (Hebei), Langfang, China\\
\{ymcui, car, wnzhang, tliu\}@ir.hit.edu.cn \\% email address must be in roman text type, not monospace or sans serif
\{ymcui, sjwang3, gphu\}@iflytek.com
}
 
\begin{document}

\maketitle

\begin{abstract}
Story Ending Prediction is a task that needs to select an appropriate ending for the given story, which requires the machine to understand the story and sometimes needs commonsense knowledge.
To tackle this task, we propose a new neural network called Diff-Net for better modeling the differences of each ending in this task. 
The proposed model could discriminate two endings in three semantic levels: contextual representation, story-aware representation, and discriminative representation. 
Experimental results on the Story Cloze Test dataset show that the proposed model siginificantly outperforms various systems by a large margin, and detailed ablation studies are given for better understanding our model. 
We also carefully examine the traditional and BERT-based models on both SCT v1.0 and v1.5 with interesting findings that may potentially help future studies.
\end{abstract}

%%%%%%%%%%%%%%%%%%
\section{Introduction}\label{introduction}
To read and comprehend human language is an important task in Artificial Intelligence (AI).
Machine Reading Comprehension (MRC) aims to comprehend the given context and answer the related questions, which is a challenging task in NLP and received extensive attention.
Owing to the availability of various large-scale datasets, we have seen rapid progress on the related neural network approaches \cite{hermann-etal-2015,kadlec-etal-2016,cui-etal-2016,wang-and-jiang-2016,seo-etal-2016,cui-acl2017-aoa,dhingra-etal-2017}. 

Story Ending Prediction is closely related to the machine reading comprehension, which aims to comprehend stories and predict the real ending, requiring world knowledge and commonsense. 
\citeauthor{mostafazadeh-etal-2016} \shortcite{mostafazadeh-etal-2016} introduced a dataset for this purpose. 
The dataset consists of a large-scale {\em unlabeled} training data consisting five-sentence commonsense stories (which forms ROCStories dataset) and {\em labeled} validation/test data, which we call Story Cloze Test (SCT). 
Figure \ref{sct-example} shows an example of SCT dataset, where the story is composed of four sentences along with one real and fake ending options, respectively.

\begin{figure}[tbp]
%\small
\centering
        \begin{tabular}{l}
        \toprule
	{\bf [Story]} \\ 
	Gina's new pencils were gone. \\ 
	And she knew a boy named Dave had taken them. \\
	She decided she would confront him to get them back. \\
	She marched to his desk and begin yelling. \\
        \midrule
	{\bf [Real Ending]} \\
	Gina was very {\bf \em angry}. \\
	{\bf [Fake Ending]} \\
	Gina was very {\bf \em calm}. \\
        \bottomrule
        \end{tabular}
\caption{\label{sct-example} An example of the Story Cloze Test dataset.}
\end{figure}

Previous works mainly focus on analyzing linguistic styles \cite{schwartz-etal-2017}, introducing external knowledge \cite{lin-etal-2017}, exploiting large-scale unlabeled training data \cite{bingning-etal-2017} etc.
In recent research, \citeauthor{cai-etal-2017} \shortcite{cai-etal-2017} propose a fully end-to-end neural network model and achieve competitive performance as previous feature-based approaches. 
\citeauthor{chaturvedi-etal-2017} \shortcite{chaturvedi-etal-2017} proposed the hidden coherence model for effectively combining several aspect models which significantly outperform traditional ensemble approaches.
Even though various efforts have been made in this task, most of them ignore the importance of better modeling endings, as two endings may both plausible, and we should pick a better one through comparisons.

To this end, we propose to address the comparisons of two endings and design an effective neural network instead of feature engineering or sophisticated reasoning rules.
As shown in Figure \ref{sct-example}, the two candidate endings are quite similar except for the last word, which is the dominant component of representing the sentiment of the whole sentence. Inspired by this, we propose to model the {\em different} part of two endings for better identifying the real ending.
We propose Diff-Net to model the differences between real and fake endings on different levels in the neural network, including contextual representation, story-aware  representation, and discriminative representation. 
The main contributions are as follows.
\begin{itemize}[leftmargin=*]
	\item We propose a novel neural network called Diff-Net for better modeling the differences of endings for the SCT task.
	\item Experimental results on the SCT test v1.0 and v1.5 dataset show that the proposed model could significantly outperform various systems by a large margin. 
	\item Detailed ablations are given, and several analyses on the SCT dataset are carried out with interesting observations that may help future studies.
\end{itemize}

%%%%%%%%%%%%%%%%%%%%%%%%%%%%%%%%%%%%%%%%%
\section{Related Works}\label{related-works}
Machine reading comprehension is a task to read and comprehend given articles and answer the questions based on them, whose subtasks include cloze-style RC, span-extraction RC, open-domain RC, commonsense RC, etc. 

CNN/DailyMail \cite{hermann-etal-2015} and Children's Book Test (CBT) \cite{hill-etal-2015} are the representative cloze-style reading comprehension datasets and various neural network models have been proposed which become cornerstones for future research, such as Attentive Reader \cite{hermann-etal-2015}, Attention Sum (AS) Reader \cite{kadlec-etal-2016}, Attention-over-Attention (AoA) Reader \cite{cui-acl2017-aoa}, Gated-Attention (GA) Reader \cite{dhingra-etal-2017}. 

As a natural extension to the cloze-style reading comprehension, \citeauthor{rajpurkar-etal-2016} \shortcite{rajpurkar-etal-2016} proposed a large-scale crowdsourced dataset SQuAD, which aims to answer the question using a span in the passage. After the release of SQuAD, various neural network models are proposed, including Match-LSTM \cite{wang-and-jiang-2016}, Bi-Directional Attention Flow (BiDAF) \cite{seo-etal-2016} etc. In recent SQuAD leaderboard, the performances of several ensemble models have already surpassed the average human performance on this dataset.

Except for the efforts on SQuAD-like datasets, some researchers are heading for another aspect of reading comprehension.
They are investigating how to incorporate external knowledge to improve the performance of reading comprehension. 
In this context, \citeauthor{mostafazadeh-etal-2016} \shortcite{mostafazadeh-etal-2016} introduced a dataset called ROCStories, which consists of five-sentence commonsense stories. To evaluate the story comprehension, they also propose evaluation sets called Story Cloze Test (SCT) aiming to comprehend given stories and predict the real ending with world knowledge and commonsense. 
This dataset is composed of two parts: one large-scale unlabeled training data and labeled validation/test data.
They also introduced several baselines for this task, and the results indicate that the current models are far from human performance. 

The related works on this dataset are mainly divided into two genres: exploiting large-scale unlabeled training data or designing a more effective model.
Different from other types of reading comprehension tasks, the training data of the SCT dataset is unlabeled, and it is hard to train a reliable supervised model (such as neural networks, etc.). 
To tackle this issue, \citeauthor{bingning-etal-2017} \shortcite{bingning-etal-2017} propose a generative adversarial network for exploiting the unlabeled training data.
Also, some researchers proposed to directly use the validation data as training data and evaluate their model performance on the test data \cite{cai-etal-2017,chaturvedi-etal-2017}.
On the other hand, \citeauthor{schwartz-etal-2017} \shortcite{schwartz-etal-2017} focus on the writing styles in this task and build a linear classifier with language models to tackle this task. 
\citeauthor{lin-etal-2017} \shortcite{lin-etal-2017} propose a multi-knowledge reasoning approach which could exploit heterogeneous knowledge.
\citeauthor{cai-etal-2017} \shortcite{cai-etal-2017} propose a fully end-to-end neural network model, and it shows competitive performance as previous precisely designed feature-based approaches. They also mentioned that by using the ending itself could give competitive performances to the ones that using story information.
\citeauthor{chaturvedi-etal-2017} \shortcite{chaturvedi-etal-2017} proposed the hidden coherence model for effectively combining several aspect feature models which outperform traditional ensemble approaches.

\begin{figure*}[htbp]
  \centering
  \includegraphics[width=0.9\textwidth]{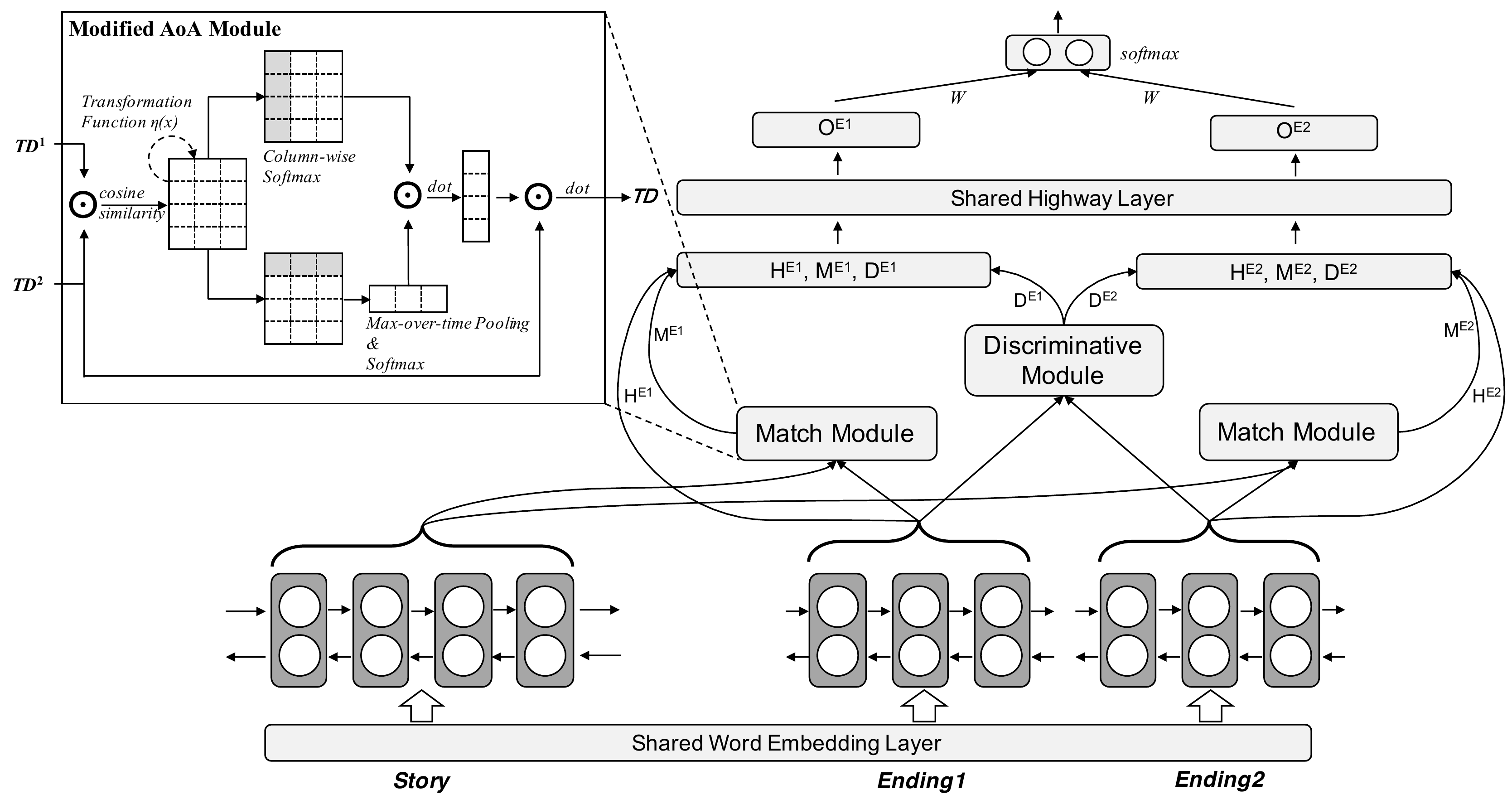}
  \caption{\label{nn-arch} Neural network architecture of the proposed Diff-Net model. Note that, several operations are omitted for simplicity. Both Match Module and Discriminative Module share the same Modified AoA Module without/with transformation function, respectively.}
\end{figure*}

Though various efforts have been made, we have found that most of the previous works neglect the importance of modeling two endings.
To this end, in this paper, we propose a fully end-to-end neural network called Diff-Net, which aims to model the differences between the real and fake endings in various semantic levels to fully discriminate them. 
We also perform detailed ablation studies to better demonstrate the importance of each component in our model.
Finally, careful analysis of traditional and BERT-based model on both SCT v1.0 and v1.5 had been carried out, which reveals how these models achieve good results on this task.

%%%%%%%%%%%%%%%%%%%%%%
\section{The Approach}\label{diffnet}
\subsection{Story Ending Prediction}
First, we give a formal definition of the story ending prediction task.
Specifically, we describe the Story Cloze Test dataset \cite{mostafazadeh-etal-2016}.
Given a four-sentence story $S=\{s_1, s_2, s_3, s_4\}$, we aim to predict real ending $\widetilde{E}$ among two candidate endings $E^i$.
In most cases, both endings are relevant to the story, but there is only one appropriate ending where sometimes need commonsense knowledge to distinguish them.
\begin{equation}
\widetilde{E} = \mathop{\arg\max}_{i=1,2}P(S, E^i)
\end{equation}

%%%%%%%%%%%%%%%%%%%%%%
\subsection{Diff-Net}
In this paper, we propose a neural network called Diff-Net to dynamically model the differences between two endings and fully utilize them for predicting the real ending. 
We discriminate two endings in three semantic aspects: contextual representation, story-aware representation, and discriminative representation.

We step in our model with three inputs.
The first one is the input of the story $S$. We do not encode each sentence individually and regard the whole story as a single long sentence.
The other two inputs are the candidate endings $E^i$, which contains the real and fake ending.
We use the Siamese network \cite{chopra-etal-2005} to encode two endings with shared weights in order to encode them freely regardless of its order of the entry.
Also, unless indicated, the weights are not shared across the story and endings.
Throughout this section, we use $t$ for representing $t$th word in the story or ending, and the super-script $S$ for the story, $E$ for the ending, which can be either $E^1$ or $E^2$.

%%%%%%%%%%%%%%%%%%%%%%
\subsubsection*{$\bullet$~~ Contextual Representation}
We first transform word in the story and two endings into one-hot representations and then convert them into continuous word embeddings with a shared pre-trained word embedding matrix $W_e$. As the dataset is relatively small, using pre-trained word embedding may mitigate the lack of data. 
Also, we add the following binary features into embedding representations of two endings.
\begin{itemize}
	\item {\bf End-End Match}: $F'_t$ is 1 if ending's token $E(t)$ appears in another ending, otherwise is 0, which is designed to highlight the same/different parts on the token level.
	\item {\bf End-Story Match}: $F''_t$ is 1 if ending's token $E(t)$ appears in the story, otherwise is 0, which is designed to highlight the relevant parts w.r.t. the story.
	\item {\bf End-Story Fuzzy Match}: We also use End-Story word-stemmed matching feature $F'''_t$ to mitigate the mismatch by the plural form, verb tense, etc., which is generated by NLTK toolkit \cite{nltk} with Porter Stemmer \cite{porter1980algorithm}.\footnote{We also tried lemmatization using such as NLTK WordNet lemmatizer but did not outperform the simple stemmer.}
\end{itemize}

This will form the enriched embeddings $E^E \in \mathbb{R}^{|E| \times (e+3)}$, where $e$ represents embedding vector size. Note that, we do not add features to the story embedding, thus $E^S \in \mathbb{R}^{|S| \times e}$.
\begin{gather}
E^S_t = W_e \cdot S(t) \\
E^E_t = [W_e \cdot E(t) ; F'_t ; F''_t ; F'''_t]
\end{gather}

After obtaining word embeddings, we further use bi-directional LSTM \cite{graves-etal-2005} to get contextual representations of the story and endings individually. 
Note that, we do not use shared LSTMs between story and endings, because their content and semantic distributions are relatively different, as indicated by \citeauthor{cai-etal-2017} \shortcite{cai-etal-2017}. 
Then we get Bi-LSTM representation of each word in the story $\overleftrightarrow{H^S} \in \mathbb{R}^{|S| \times 2h}$ and endings $\overleftrightarrow{H^E}  \in \mathbb{R}^{|E| \times 2h}$ by concatenating forward and backward LSTM hidden states in each time step, where $h$ represents hidden vector size for one direction. We take the ending representation $\overleftrightarrow{H^E}$ as an example, as shown in Equation 4 and 5.
\begin{gather}
\overleftrightarrow{H^E} = [\overrightarrow{\mathbf{LSTM}}(E^E) ~~;~~ \overleftarrow{\mathbf{LSTM}}(E^E)] 
\end{gather}
Then we obtain the flattened representations of endings $H^{E} \in \mathbb{R}^{2h}$ by applying max-over-time pooling on the Bi-LSTM representations $\overleftrightarrow{H^E}$.
\begin{gather}
H^{E} = \mathop{MaxPooling}_{t=1,...,N}(\overleftrightarrow{H^E_t})
\end{gather}

%%%%%%%%%%%%%%%%%%%%%%
\subsubsection*{$\bullet$~~ Match Module}
In this module, we are focusing on calculating the similarity between the story and each ending to measure the coherency between them using the contextual representations.

To enrich the representation of the story and endings, we further apply a fully-connected layer with SELU activation (denoted as $\sigma$) \cite{klambauer-etal-2017} to every time step, forming enriched representations $R^S \in \mathbb{R}^{|S| \times h}$ and $R^E \in \mathbb{R}^{|E| \times h}$ respectively. Note that, we halve the hidden size for compact representations.
\begin{gather}
R^S_t = \sigma(W_{rs} \cdot \overleftrightarrow{H^S_t} + b_{rs}) \\
R^E_t = \sigma(W_{re} \cdot \overleftrightarrow{H^E_t} + b_{re})
\end{gather}

To calculate the similarity between the story and ending, we adopt the Attention-over-Attention (AoA) mechanism \cite{cui-acl2017-aoa} for its strong empirical performance. 
In this paper, we make some modifications to the AoA mechanism to get story-aware ending representations $M^E$.
First, we replace the dot calculation with cosine similarity. 
Second, we adopt max-pooling instead of average-pooling. 
Third, we introduce a transformation function for the similarity matrix. 
We will give a formal definition of the proposed modified AoA module, as we will reuse this module in the following part. 
Note that there is no weight matrix applied in the AoA mechanism.
The general formulation is given in Algorithm \ref{alg:modified-aoa}.
We will introduce how the modified AoA applied in Diff-Net.

Given the story representation $R^S$ and one ending representation $R^E$, firstly, we calculate cosine similarity between one story word and one ending word forming a similarity matrix.
The transformation function is applied when the $\eta$ function is provided. 
Then we will follow the original attention-over-attention calculation. First, we apply softmax function to each column of similarity matrix to obtain story-level attention $\alpha$ w.r.t. each ending word. Then, we transpose the similarity matrix and do max-over-time pooling and apply softmax function to get a unified ending-level attention $\beta$, indicating the importance of each ending word w.r.t. the story. Finally, the attention-over-attention $\gamma$ is obtained by calculating the weighted sum of $\alpha$ and $\beta$. In order to get story-aware representation $M^E$, we simply calculate the weighted sum of $R^E$ given attention value $\gamma$. 
A visual depiction of the modified AoA mechanism can be seen in Figure \ref{nn-arch}.
Altogether, we can get story-aware representation for each ending $M^{E^1}$, $M^{E^2}  \in \mathbb{R}^{h}$. 
\begin{gather}
M^{E^1} = \mathbf{AoA}(R^S, R^{E^1}, \text{None})  \\
M^{E^2} = \mathbf{AoA}(R^S, R^{E^2}, \text{None}) 
\end{gather}

% alias for REQUIRE / ENSURE
\renewcommand{\algorithmicrequire}{\textbf{Input:}} 
\renewcommand{\algorithmicensure}{\textbf{Output:}}

\begin{algorithm}[tb]
%\small
\caption{Modified Attention-over-Attention.}   
\label{alg:modified-aoa}   
\begin{algorithmic}[1] %这个1 表示每一行都显示数字  
\REQUIRE ~~\\ %算法的输入参数：Input  
Time-Distributed representation {\em TD}$^1$ \\
Time-Distributed representation {\em TD}$^2$ \\
Transformation Function $\eta(x)$, default None \\
\ENSURE {\em TD}$^1$-aware {\em TD}$^2$ representation {\em TD} 
\STATE Calculate similarity matrix between {\em TD}$^1_i$ and {\em TD}$^2_j$: \\$sim(i,j) = cos<${\em TD}$^1_i, ${\em TD}$^2_j>$ \\
\STATE Skip if $\eta$ is None, else apply transformation function $\eta$: \\$sim(i,j) = \eta(sim(i,j))$ \\
\STATE Attention for each {\em TD}$^2_j$: \\
$\alpha(t)=softmax(sim(1,t), ..., sim(M,t))$ ; \\
$\alpha = [\alpha(1); ...; \alpha(N)]$ \\
\STATE Transpose similarity matrix: $sim^T = transpose(sim)$
\STATE Max-over-time pooling: $sim_{M}=max\text{-}pool(sim^T)$	% on $sim^T$
\STATE Softmax over $sim_{M}$: $\beta = softmax(sim_{M})$
\STATE Attention-over-Attention: $\gamma = softmax(\alpha^{T} \beta) $
\STATE {\em TD}$^1$-aware {\em TD}$^2$ representation: {\em TD}$ = (${\em TD}${^2})^T \gamma$
\RETURN {\em TD}
\end{algorithmic}  
\end{algorithm}

%%%%%%%%%%%%%%%%%%%%%
\subsubsection*{$\bullet$~~ Discriminative Module}
To further characterize each ending, we will proceed with modeling {\em differences} between two endings. 
We still use modified AoA, except for enabling the transformation function $\eta$ (Algorithm 1, Line 2) to get the discriminative ending representations $D^{E^1}$, $D^{E^2}\in\mathbb{R}^{h}$. 
\begin{gather}
D^{E^1} = \mathbf{AoA}(R^{E^2}, R^{E^1}, \eta) \\
D^{E^2} = \mathbf{AoA}(R^{E^1}, R^{E^2}, \eta)
\end{gather}

The transformation function $\eta$ is designed to depict {\em dissimilarity} between two endings, which is important for identifying the real ending.
As the input of the transformation function is the cosine similarity $x$, we simply apply $-x$ to measure the dissimilarity between two endings.

%%%%%%%%%%%%%%%%%%%%%%
\subsubsection*{$\bullet$~~ Final Output}
After obtaining various ending representations, we will use them to predict the final probability of being the real ending.
We concatenate three ending representations from different levels: contextual representation $H^E$, story-aware representation $M^E$, discriminative representation $D^E$, and feed them into a highway network \cite{srivastava-etal-2015} with SELU activation on the output to get a hybrid representation.
Finally, we squash the output of the highway network to a single scalar score. 
We concatenate two ending scores and use the softmax function for final prediction.
\begin{gather}
O^E =  \sigma(\mathbf{Highway}([H^E; M^E; D^E])) \\
P(S, E^i) = \mathbf{softmax}([W O^{E^{1}}; W O^{E^{2}}])
\end{gather}

\subsubsection*{$\bullet$~~ Training Objective}
To train our model, we simply minimize the categorical cross entropy loss between the predictions and ground truths.
Besides, we also add additional cosine similarity loss on the highway representations of two endings $O^E$ to minimize the similarity of two representations in the latent semantic space. We have found that adding additional cosine similarity loss could further improve the performance of the model and stabllize the results as well, which will be demonstrated in the experiment section.
\begin{gather}
\mathcal{L} = \mathcal{L}_{cce} + \mathcal{L}_{cosine} \\
\mathcal{L}_{cosine} = \cos<O^{E^{1}}, O^{E^{2}}>
\end{gather}

\subsection{Incorporate BERT in Diff-Net}
Bidirectional Encoder Representation from Transformers (BERT) \cite{devlin2018bert} has shown marvelous performance in various NLP tasks. 
In order to strengthen the experimental results and demonstrate the effectiveness of proposed modules, we use two ways to incorporate BERT in our model to strengthen the baseline systems.
\begin{itemize}[leftmargin=*]
	\item {\bf BERT as Feature}: A straightforward way is to use BERT as a feature generator for the text. We concatenate flattened BERT representation, which is the average pooling result of the last hidden layer of BERT, for the highway input in Equation 11. Note that, in this case, the weights in BERT will not be updated during training phase.
	\begin{equation}
	O^E =  \sigma(\mathbf{Highway}([H^E; M^E; D^E; \text{BERT}^E]))
	\end{equation}
	\item {\bf Fine-tuning BERT}: Another way is to fine-tune the BERT on SCT task directly. In this setting, we obtain two BERT representation with shared BERT weights for {\tt[end1, story]} and {\tt[end2, story]} as two ending representations and feed into the Discriminative Module. The final predictions are made by concatenating BERT representation and discriminative representation, where Equation 11 can be rewritten as follows. 
	The final predictions are similarly made as Equation 12.
	\begin{gather}
	O^E =  \sigma(\mathbf{Highway}([D^E; \text{FT-BERT}^E]))
	\end{gather}
\end{itemize}

%%%%%%%%%%%%%%%%%%%%%%%%%%%%%%%%%%%%%%%%%
\section{Experiments}\label{experiments}        
\subsection{Experimental Setups}
We evaluate our approach on the Story Cloze Test dataset \cite{mostafazadeh-etal-2016}, which consists of 1,871 validation and test set, respectively.
Following previous works \cite{cai-etal-2017,chaturvedi-etal-2017,schwartz-etal-2017}, we take the validation set for training and evaluate the performance on the test set. 
Note that, unlike using the scheme of `pre-training then fine-tuning' \cite{liu-etal-2017}, we did not exploit 50k unlabeled ROCStories training data to do task-specific pre-training.
The basic settings of the proposed model are listed below in detail.
\begin{itemize}
	\item {\bf Embedding Layer}: The embedding weights are initialized by the pre-trained GloVe vectors (840B version, 300 dimension) \cite{pennington-etal-2014} and updated during the training phase. Note that, there are also three additional binary features concatenated to the Glove embedding.
	\item {\bf Hidden Layer}: We use Bi-LSTM with 200 dimension for each direction (400 in total), and 200 for the enriched representation $R$.
	\item {\bf Regularization}: We apply $l_2$-regularization of 0.001 on the embedding weights and a dropout \cite{srivastava-etal-2014} rate of 0.5 after the embedding layer. 
	\item {\bf Optimization}: We use \textsc{Adam} for weight updating \cite{kingma2014adam} with an initial learning rate of 0.001, and decaying learning rate at each epoch by a factor of 0.8. Also we clipped the $l_2$-norm of gradient to 5 \cite{pascanu-etal-2013} to avoid gradient explosion. The batch size is set to 32.
\end{itemize}

The results are reported with the best model, and the ensemble models are trained on the different random seeds, and we use the majority voting approach.
Our implementation is based on Keras \cite{chollet2015keras} and TensorFlow \cite{abadi2016tensorflow}. 
Traditional neural models are trained on NVIDIA Tesla V100 GPU. BERT related models are trained on a single TPU v2, which has 64G HBM.
For the models that using BERT, specifically, we use uncased BERT-large pre-trained model, which has 1024 hidden dimension and 24 transformer layers.

%%%%%%%%%%%%%%%%
\subsection{Results on SCT v1.0}
The overall results are shown in Table \ref{overall-results}. 
As we can see that our Diff-Net achieves an accuracy of 77.8, which outperforms previous system by an absolute gain of 2.6\% compared to \citeauthor{schwartz-etal-2017} \shortcite{schwartz-etal-2017}, demonstrating the effectiveness of the proposed model.
Note that, the best previous single model by \citeauthor{schwartz-etal-2017} \shortcite{schwartz-etal-2017} adds additional RNNLM trained on the large-scale training set, while we do not exploit any knowledge from the training data. 

When it comes to the ensemble model, we can see that the proposed model also brings significant improvements compared to the traditional majority voting or soft voting approaches. 
Also, our model surpasses the previous hybrid system by the Hidden Coherence Model (HCM) \cite{chaturvedi-etal-2017} with 1.4\% improvements in accuracy.

By incorporating the BERT representation, our model could further boost up to 84.3.
However, the fine-tuned BERT model itself could beat all previous single and ensemble systems with an accuracy of 89.2.
After incorporating Discriminative Module, BERT+Diff-Net could give another improvements over fine-tuned BERT with an accuracy of 90.1, demonstrating its effectiveness.

\begin{table}[tp]
\small
\begin{center}
\begin{tabular}{lc}
\toprule
{\bf System} & {\bf Test } \\
\midrule
	DSSM \cite{mostafazadeh-etal-2016} 						& 58.5 \\
	Conditional GAN \cite{bingning-etal-2017} 					& 60.9 \\
	Heterogeneous Knowledge \cite{lin-etal-2017} 				& 67.0 \\
	Linguistic Style \cite{schwartz-etal-2017} 					& 72.4 \\ 
	Hier, EndingOnly \cite{cai-etal-2017} 						& 72.5 \\
	Logistic Regression \cite{chaturvedi-etal-2017} 				& 74.4 \\
	Hier, EncPlotEnd, Att \cite{cai-etal-2017}					& 74.7 \\
	Linguistic Style + RNNLM \cite{schwartz-etal-2017} 			& 75.2 \\
	\bf Diff-Net											& {\bf 77.8} \\
	\bf Diff-Net + BERT										& {\bf \underline{84.3}} \\	
	\midrule
	Majority Voting (ensemble)$^*$ 							& 69.5 \\
	Soft Voting (ensemble)$^*$								& 75.1 \\
	HCM (3-aspects)$^*$									& 77.6 \\	
	\bf Diff-Net (ensemble) 									& {\bf 79.0} \\
	\midrule \midrule
	\em{Fine-tuned Pre-trained Systems}\\
	Transformer LM \cite{openai-gpt}							& \underline{86.5} \\
	\bf BERT												& {\bf \underline{89.2}} \\ 
	\bf BERT + Diff-Net										& {\bf \underline{90.1}} \\
\bottomrule
\end{tabular}
\end{center}
\caption{\label{overall-results} Overall results (accuracy) on Story Cloze Test v1.0 dataset. Our results are depicted in bold face (verified by the significant test with $p<0.05$). Result with underline means it utilizes pre-trained model. * indicates results from \citeauthor{chaturvedi-etal-2017} \shortcite{chaturvedi-etal-2017}.   }
\end{table}

%%%%%%%%%%%%%%%%
\subsection{Results on SCT v1.5}
SCT v1.5 \cite{sharma-etal-2018-tackling} is an evolved version of SCT v1.0, which alleviate the ending bias in the original dataset.
Similar to v1.0 settings, we use the development set (1,571 samples) for training.
Different from v1.0, the test set v1.5 is hidden to the public, and we submit our best performing model on test v1.0 to get the final results.
From Table \ref{sct15-results}, as we can see that, our model could also give consistent improvements over fine-tuned BERT system, demonstrating the effectiveness of the proposed model.

\begin{table}[t]
\small
\begin{center}
\begin{tabular}{p{3.5cm} cc}
\toprule
{\bf System} 					& {\bf Test v1.0} & {\bf Test v1.5} \\
\midrule
word2vec$^\dag$					& 53.9 & 59.4 \\
EndingReg$^\dag$					& 71.5 & 64.4 \\
cogcomp$^\dag$					& 77.6 & 60.8 \\
\midrule
\bf BERT 						& 88.7 & 80.7 \\
\bf BERT + Diff-Net 			& 90.1 & 82.0 \\
\bottomrule
\end{tabular}
\end{center}
\caption{\label{sct15-results} Results on SCT v1.5 dataset. $^\dag$ indicates the results from \citeauthor{sharma-etal-2018-tackling} \shortcite{sharma-etal-2018-tackling}.   }
\end{table}

%%%%%%%%%%%%%%%%
\subsection{Ablation Studies}
In the following parts, we will carry out detailed ablation studies to better understand our model.
In order to avoid training instability on small data and non-determinism of TensorFlow under GPU or TPU, except for reporting the best accuracy, we also report the average and standard deviation through 10 independent runs.\footnote{The average accuracy and standard deviation are shown in the brackets. Note that, due to the Non-Gaussian distribution of the results, {\tt average + stdev $\ne$ max}.}
We ablate on Diff-Net without BERT to eliminate the effect by its powerful empirical performance to solely evaluate on our model.
We will analyze the results based on the average score and its standard deviation, which is statistically stable and reliable.
We first begin with the general components in our Diff-Net, and the results are shown in Table \ref{ablation-general}.

\begin{table}[htbp]
\small
\begin{center}
\begin{tabular}{p{4cm} c}
\toprule
{\bf System} 					& {\bf Test v1.0 Accuracy} \\
\midrule
	\bf Diff-Net 					& \bf 77.8 ($77.60 \pm 0.12$) \\
	L1: SELU $\rightarrow$ Tanh 		& 77.2 ($77.04 \pm 0.06$) \\
	L2: w/o cosine loss				& 77.2 ($77.02 \pm 0.20$) \\
	L3: w/o modified AoA 			& 77.0 ($76.93 \pm 0.07$) \\
	L4: w/o match module 			& 77.1 ($76.85 \pm 0.20$) \\
	L5: w/o discriminative module 		& 76.9 ($76.75 \pm 0.21$) \\
	L6: AoA $\rightarrow$ dot product	& 76.7 ($76.58 \pm 0.11$) \\
	L7: w/o all binary features 		& 76.0 ($75.78 \pm 0.12$) \\
\bottomrule
\end{tabular}
\end{center}
\caption{\label{ablation-general} Ablations on Diff-Net. The results are ordered by the descending average score. For reference simplicity, we label each experiment with L1 to L7.}
    \end{table}    

When compared to the original AoA mechanism (L3), the modified AoA gives 0.67\% improvements on average, indicating that the modified AoA is more powerful in the context of our model. Also, it demonstrates that the max-pooling is relatively superior at filtering noise and choose the most representative values in the vectors. If we replace the AoA mechanism to the simple dot product (L6), there is a significant drop by near 1\%, suggesting that the AoA mechanism is helpful in precisely calculating attentions.
By discarding the match module (L4), we see a significant drop in performance by 0.75\%, while without the diff module (L5) gives an even lower score.
This demonstrates that retrieving relevant information from the story as well as discriminating two endings are both important in this task, and combining these components yields further improvements.
If we remove the cosine loss in training objective (L2), there is a slight drop in the performance as well as brings bigger fluctuation in results, indicating that adding additional loss could stabilize the experimental results, which demonstrates that separating latent semantic distance between two endings are helpful in this task.
Also, as we can see that, without using any binary features (L7) will bring a significant drop in performance, which demonstrates that the matching features will help the model better recognize the alignment between the story and endings in the traditional models.

Recall that we have added three binary features to enhanced word embedding representation of the endings, especially when the training data is not enough. 
The ablation results are given in Table \ref{ablation-feature}.
Among three binary features, the most useful one is the end-story fuzzy matching feature, in the meantime, it could give slight improvements over the end-story non-fuzzy matching. The end-end matching feature brings moderate improvements which could be a remedy for providing mutual information of the endings.

        \begin{table}[tp]
        \begin{center}
        \begin{tabular}{lc}
        \toprule
        {\bf System} & {\bf Test v1.0 Accuracy} \\
        \midrule
	Diff-Net (all features)~~~~					& 77.8 \small($77.60 \pm 0.12$) \\
	- w/o E-E Match 						& 77.2 \small($77.05 \pm 0.11$) \\
	- w/o E-S Match 						& 77.0 \small($76.83 \pm 0.12$) \\
	- w/o E-S Fuzzy Match 					& 76.9 \small($76.78 \pm 0.11$) \\
	- w/o all features 						& 76.0 \small($75.78 \pm 0.12$) \\
        \bottomrule
        \end{tabular}
        \end{center}
        \caption{\label{ablation-feature} Ablations on using different binary features in word embedding (E: ending, S: story).}
        \end{table}

%%%%%%%%%%%%%%%%%%%%%%%%%%%%%%%%
\section{Discussion}
While BERT, as well as our modifications, brings good performance on this task, there are still several questions that remain unclear.
\begin{itemize}[leftmargin=*]
	\item[-] {\em Does models truly {\bf\em{understand or comprehend}} the story?}
	\item[-] {\em Is Story Cloze Test data (both v1.0 and v1.5) suitable for evaluating {\bf\em{story comprehension}}?}
	\item[-] {\em Except for the objective metric, in which aspects does BERT improve than the traditional neural networks?}
\end{itemize}

To investigate the questions above, we conduct comprehensive quantitative analyses to examine both models and datasets. 
We have an intuition that the last part of the story is critical for predicting the real ending. 
To verify this assumption, we discard each sentence in the story to see which sentence is of the most help in this task. 
Also, we set an experiment that does not use the story at all to see how much gain can we obtain by using the story information, to test the ability of {\em story comprehension}.
The quantitative analysis results are shown in Table \ref{ablation-context}, and we get some unanimous and interesting observations.

\begin{table}[tbp]
\small
\begin{center}
\begin{tabular}{p{1.9cm} l l l}
\toprule
\multirow{2}{*}{\bf Settings} 	& {\centering \bf Diff-Net} & \multicolumn{2}{c}{\centering \bf BERT+Diff-Net} \\
							& {\bf Test v1.0} & {\bf Test v1.0} & {\bf Test v1.5}  \\
\midrule
\bf Entire Story 		 		& \bf 77.8 & \bf 90.1 & \bf 82.0 \\
- w/o 1st sent. 				& 76.6 {\small(-1.2)} & 90.0 {\small(-0.1)} & 81.9 {\small(-0.1)} \\
- w/o 2nd sent. 				& 77.2 {\small(-0.6)} & 90.0 {\small(-0.1)} &  82.0 {\small(+0.0)} \\	
- w/o 3rd sent. 				& 77.2 {\small(-0.6)} & 89.6 {\small(-0.5) }& 82.2 {\small(+0.2)} \\
- w/o 4th sent. 				& 76.5 {\small(-1.3)} & 83.6 {\small(-6.5)} &  74.9 {\small(-7.0)} \\
\midrule
Ending only		  			& 75.9 {\small(-1.9)} & 80.6 {\small(-9.5)} & 69.1 {\small(-12.8)} \\ 
Reverse story					& 77.5 {\small(-0.3)} & 89.3 {\small(-0.8)} & 81.4 {\small(-0.5)} \\
Random order		 			& 77.6 {\small(-0.2)} & 89.3 {\small(-0.8)} & 78.7 {\small(-3.2)} \\
\bottomrule
\end{tabular}
\end{center}
\caption{\label{ablation-context} Quantitative analysis on using different proportion and sentence order of the story. The performance gap compared to the baseline is depicted in the bracket.}
\end{table}

Firstly, when we discard each sentence in the story, except for the last sentence in the story, the other sentences seem to provide little help in this task regardless of models and datasets. 
In most situations, the first sentence is providing the background information and topic of the story, and we can see that it helps in finding the correct ending only in the traditional neural network model (w/o BERT). However, when it comes to BERT-based models, there is only little variance regardless of test v1.0 or v1.5.
To our surprise, in test v1.5, though it was an evolved version of SCT, removing the second or third sentence in the story could weirdly {\em improve} overall performance, which was not expected.
We suspect that the middle sentences are not the final state of the story, thus have little impact on the ending.
Lastly, when removing the last sentence, the performance of all models in all sets decrease dramatically, which indicates that it is the most important in the story and provides key clues for predicting the real ending. 
That is, the {\em last-sentence bias} still exists in SCT v1.5.
The following example shows the last-sentence bias in this task, where we could easily pick the real ending by only looking at the last sentence in the story.

\begin{figure}[h]
%\small
\centering
        \begin{tabular}{l}
        \toprule
	{\bf [Story]} \\
	Janet worked hard to train for her wrestling meet. \\ 
	When she got there her opponent seemed game. \\
	They both tried their hardest. \\ 
	\underline{It ended in a tie.}  \\
	{\bf [Real Ending]} \\
	Janet was content with the result.  \\
	{\bf [Fake Ending]} \\
	Janet won the first place trophy. \\
        \bottomrule
        \end{tabular}
\caption{\label{discussion-sample} An example of last-sentence bias issue. By only looking at the word `tie' in the last story sentence, we can easily pick the real ending, as word `won' in fake ending raises contradiction to the story.}
\end{figure}

Secondly, in Diff-Net, there is only a 1.9\% decrease in the system performance without the presence of the story (ending only), indicating that the story does help in choosing the real ending, but the improvement is quite moderate.
However, in BERT+Diff-Net, though the baseline increases a lot, as we can see that there is about 10\% to 12\% drop without the story in the test v1.0 and v1.5 data. 
This suggests that: 1) the traditional models focus less on the story and ending itself plays a key role. 
2) The BERT-based model is better than traditional models in finding relations between the story and ending, as the input sequence is the concatenation by them and be fed into very deep transformer layers with self-attention mechanism.

Thirdly, to our surprise, reversing the story sentences or even randomly placing these sentences do not show a significant drop in the performance, which suggests that the order of the event sequence does not affect much in identifying the real ending in these datasets.
Nonetheless, there is a significant drop in the test v1.5 compared to the counterparts, which demonstrate the test v1.5 does improve the evaluation on the narrative order of the story, but not that salient (only -3.2\% in the accuracy). 
While \citeauthor{chen-etal-2016-order} \shortcite{chen-etal-2016-order} discussed the importance of sentence ordering with respect to the story coherency, according to the results above, it seems not to be a crucial component in current {\em story comprehension} dataset.

Also, it can be inferred that current models are treating the sentences in the story as {\em discrete clues} rather than a temporal event sequence. Thus, we suspect the effect of using script knowledge for helping this task is quite limited, and we would investigate this in the future.

%%%%%%%%%%%%%%%%%%%%%%%%%%%%%%%%%%%%%%%%%
\section{Conclusion}\label{conclusion}
In this paper, we proposed a novel neural network model called Diff-Net to tackle the story ending prediction task.
Our model could dynamically model the ending differences in three aspects and retrieve relevant information from the story.
Also, we propose to use additional cosine objective function to separate the latent semantic distance between two ending representations.
Experimental results on SCT v1.0 and v1.5 show that the proposed model could bring significant improvements over traditional neural baselines and BERT baselines.
Except for the proposed model, we also carried out quantitative analyses on both traditional and BERT models and concluded that there is still a long way to go to achieve actual story comprehension.

As we indicated, the order of the story sentences does not affect the final performance much, in the future, we are going to verify our assumptions by introducing script knowledge to see if this could help in identifying real ending. 
Also, we would like to investigate the potential usage of unlabeled training data, such as training pre-trained models or constructing knowledge base for this task.

%%%%%%%%%%%%%%%%%%%%%%%%%%%%%%%%%%%%%%%%%
\section{Acknowledgments}\label{ack}
We would like to thank all anonymous reviewers and senior program members for their thorough reviewing and providing constructive comments to improve our paper. 
The first author was partially supported by the Google TensorFlow Research Cloud (TFRC) program for Cloud TPU access.
This work was supported by the National Natural Science Foundation of China (NSFC) via grant 61976072, 61632011, and 61772153.

\fontsize{9.5pt}{10.5pt} \selectfont 
\bibliography{AAAI-CuiY.7157}
\bibliographystyle{aaai}

\end{document}